\documentclass[runningheads]{llncs}
\usepackage{graphicx}
\usepackage{amsmath,amssymb} 
\usepackage{color}

\begin{document}
\title{Toward Characteristic-Preserving Image-based Virtual Try-On Network} 

\titlerunning{CP-VTON}
%
\author{Bochao Wang\inst{1,2}\and
Huabin Zheng\inst{1,2} \and
Xiaodan Liang\inst{1}\thanks{The corresponding author is Xiaodan Liang} \and
Yimin Chen\inst{2} \and
Liang Lin\inst{1,2} \and
Meng Yang\inst{1}}
%
\authorrunning{Bochao Wang et al.}
%

\institute{Sun Yat-sen University, China \and
SenseTime Group Limited \\
\email{\{wangboch,zhhuab\}@mail2.sysu.edu.cn,xdliang328@gmail.com, chenyimin@sensetime.com, linliang@ieee.org, yangm6@mail.sysu.edu.cn}}
\maketitle              
\begin{abstract}
Image-based virtual try-on systems for fitting a new in-shop clothes into a person image have attracted increasing research attention, yet is still challenging.
A desirable pipeline should not only transform the target clothes into the most fitting shape seamlessly but also preserve well the clothes identity in the generated image, that is, the key characteristics (e.g. texture, logo, embroidery) that depict the original clothes.
However, previous image-conditioned generation works fail to meet these critical requirements towards the plausible virtual try-on performance since they fail to handle large spatial misalignment between the input image and target clothes.
Prior work explicitly tackled spatial deformation using shape context matching, but failed to preserve clothing details due to its coarse-to-fine strategy.
In this work, we propose a new fully-learnable Characteristic-Preserving Virtual Try-On Network (CP-VTON) for addressing all real-world challenges in this task. 
First, CP-VTON learns a thin-plate spline transformation for transforming the in-shop clothes into fitting the body shape of the target person via a new Geometric Matching Module (GMM) rather than computing correspondences of interest points as prior works did.
Second, to alleviate boundary artifacts of warped clothes and make the results more realistic, we employ a Try-On Module that learns a composition mask to integrate the warped clothes and the rendered image to ensure smoothness.
Extensive experiments on a fashion dataset demonstrate our CP-VTON achieves the state-of-the-art virtual try-on performance both qualitatively and quantitatively.
Code is available at \url{https://github.com/sergeywong/cp-vton}.

\end{abstract}

%
%
\section{Introduction}
Online apparel shopping has huge commercial advantages compared to traditional shopping(e.g. time, choice, price) but lacks physical apprehension. 
To create a shopping environment close to reality, virtual try-on technology has attracted a lot of interests recently by delivering product information similar to that obtained from direct product examination.
It allows users to experience themselves wearing different clothes without efforts of changing them physically.
This helps users to quickly judge whether they like a garment or not and make buying decisions, and improves sales efficiency of retailers.
The traditional pipeline is to use computer graphics to build 3D models and render the output images since graphics methods provide precise control of geometric transformations and physical constraints.
But these approaches require plenty of manual labor or expensive devices to collect necessary information for building 3D models and massive computations.

\begin{figure*}[t]
\centering
\includegraphics[scale=0.16]{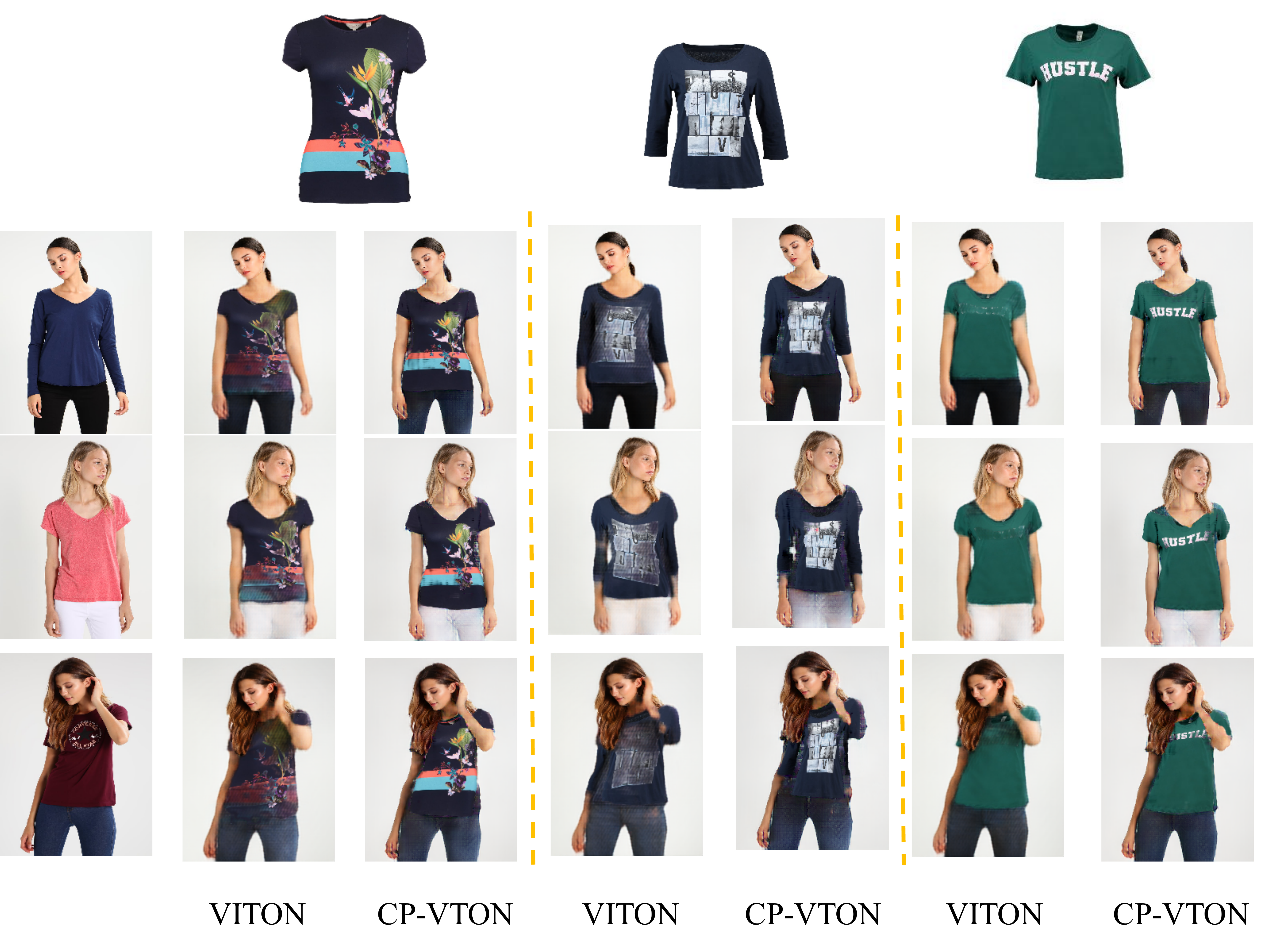}\vspace{-5mm}
\caption{The proposed CP-VTON can generate more realistic image-based virtual try-on results that preserve well key characteristics of the in-shop clothes, compared to the state-of-the-art VITON~\cite{han2017viton}.}\vspace{-8mm}
\label{fig:visualization}
\end{figure*}

More recently, the image-based virtual try-on system~\cite{han2017viton} without resorting to 3D information, provides a more economical solution and shows promising results by reformulating it as a conditional image generation problem.
Given two images, one of a person and the other of an in-shop clothes, such pipeline aims to synthesize a new image that meets the following requirements:
a) the person is dressed in the new clothes; 
b) the original body shape and pose are retained;
c) the clothing product with high-fidelity is warped smoothly and seamlessly connected with other parts;
d) the characteristics of clothing product, such as texture, logo and text, are well preserved, without any noticeable artifacts and distortions.
Current research and advances in conditional image generation (e.g. image-to-image translation~\cite{isola2017image,zhu2017unpaired,choi2017stargan,wang2017high,liang2017generative,deng2017structured}) make it seem to be a natural approach of facilitating this problem. 
Besides the common pixel-to-pixel losses (e.g. L1 or L2 losses) and perceptual loss~\cite{johnson2016perceptual}, an adversarial loss~\cite{isola2017image} is used to alleviate the blurry issue in some degree, but still misses critical details.
Furthermore, these methods can only handle the task with roughly aligned input-output pairs and fail to deal with large transformation cases.
Such limitations hinder their application on this challenging virtual try-on task in the wild.
One reason is the poor capability in preserving details when facing large geometric changes, e.g. conditioned on unaligned image~\cite{ma2017pose}.
The best practice in image-conditional virtual try-on is still a two-stage pipeline VITON~\cite{han2017viton}. 
But their performances are far from the plausible and desired generation, as illustrated in Fig.~\ref{fig:visualization}. 
We argue that the main reason lies in the imperfect shape-context matching for aligning clothes and body shape, and the inferior appearance merging strategy.

To address the aforementioned challenges, we present a new image-based method that successfully achieves the plausible try-on image syntheses while preserving cloth characteristics, such as texture, logo, text and so on, named as Characteristic-Preserving Image-based Virtual Try-On Network (CP-VTON). 
In particular, distinguished from the hand-crafted shape context matching, we propose a new learnable thin-plate spline transformation via a tailored convolutional neural network in order to align well the in-shop clothes with the target person. 
The network parameters are trained from paired images of in-shop clothes and a wearer, without the need of any explicit correspondences of interest points.
Second, our model takes the aligned clothes and clothing-agnostic yet descriptive person representation proposed in~\cite{han2017viton} as inputs, and generates a pose-coherent image and a composition mask which indicates the details of aligned clothes kept in the synthesized image.
The composition mask tends to utilize the information of aligned clothes and balances the smoothness of the synthesized image.
Extensive experiments show that the proposed model handles well the large shape and pose transformations and achieves the state-of-art results on the dataset collected by Han et al.~\cite{han2017viton} in the image-based virtual try-on task.

Our contributions can be summarized as follows:
\begin{itemize}
    \item We propose a new Characteristic-Preserving image-based Virtual Try-On Network (CP-VTON) that addresses the characteristic preserving issue when facing large spatial deformation challenge in the realistic virtual try-on task.
    \item Different from the hand-crafted shape context matching, our CP-VTON incorporates a full learnable thin-plate spline transformation via a new Geometric Matching Module to obtain more robust and powerful alignment.
    \item Given aligned images, a new Try-On Module is performed to dynamically merge rendered results and warped results.
	\item Significant superior performances in image-based virtual try-on task achieved by our CP-VTON have been extensively demonstrated by experiments on the dataset collected by Han et al.~\cite{han2017viton}.
\end{itemize}

%
%
\section{Related Work}
\subsection{Image synthesis}
Generative adversarial networks(GANs)~\cite{goodfellow2014generative} aim to model the real image distribution by forcing the generated samples to be indistinguishable from the real images.
Conditional generative adversarial networks(cGANs) have shown impressive results on  image-to-image translation, whose goal is to translate an input image from one domain to another domain ~\cite{isola2017image,zhu2017unpaired,choi2017stargan,wang2017high,li2017perceptual,liang2017dual,yang2018unsupervised}.
Compared L1/L2 loss, which often leads to blurry images, the adversarial loss has become a popular choice for many image-to-image tasks.
Recently, Chen and Koltun~\cite{chen2017photographic} suggest that the adversarial loss might be unstable for high-resolution image generation.
We find the adversarial loss has little improvement in our model.
In image-to-image translation tasks, there exists an implicit assumption that the input and output are roughly aligned with each other and they represent the same underlying structure.
However, most of these methods have some problems when dealing with large spatial deformations between the conditioned image and the target one.
Most of image-to image translation tasks conditioned on unaligned images~\cite{han2017viton,ma2017pose,zhao2017multi}, adopt a coarse-to-fine manner to enhance the quality of final results.
To address the misalignment of conditioned images, Siarohit et al.~\cite{siarohin2017deformable} introduced a deformable skip connections in GAN, using the correspondences of the pose points.
VITON~\cite{han2017viton} computes shape context thin-plate spline(TPS) transofrmation~\cite{belongie2002shape} between the mask of in-shop clothes and the predicted foreground mask.
Shape context is a hand-craft feature for shape and the matching of two shapes is time-consumed. 
Besides, the computed TPS transoformations are vulnerable to the predicted mask.
Inspired by Rocco et al.~\cite{rocco2017convolutional}, we design a convolutional neural network(CNN) to estimate a TPS transformation between in-shop clothes and the target image without any explicit correspondences of interest points.

\subsection{Person Image generation}
Lassner et al.~\cite{lassner2017generative} introduced a generative model that can generate human parsing~\cite{gong2017look} maps and translate them into persons in clothing.
But it is not clear how to control the generated fashion items.
Zhao et al.~\cite{zhao2017multi} addressed a problem of generating multi-view clothing images based on a given clothing image of a certain view.
PG2~\cite{ma2017pose} synthesizes the person images in arbitrary pose, which explicitly uses the target pose as a condition.
Siarohit et al.~\cite{siarohin2017deformable} dealt the same task as PG2, but using the correspondences between the target pose and the pose of conditional image.
The generated fashion items in \cite{zhao2017multi,ma2017pose,siarohin2017deformable}, kept consistent with that of the conditional images.
FashionGAN~\cite{zhu2017your} changed the fashion items on a person and generated new outfits by text descriptions.
The goal of virtual try-on is to synthesize a photo-realistic new image with a new piece of clothing product, while leaving out effects of the old one.
Yoo te al.~\cite{yoo2016pixel} generated in shop clothes conditioned on a person in clothing, rather than the reverse.

\subsection{Virtual Try-on System}
Most virtual try-on works are based on graphics models.
Sekine et al.~\cite{sekine2014virtual} introduced a virtual fitting system that captures 3D measurements of body shape.
Chen et al.~\cite{chen2016synthesizing} used a SCAPE~\cite{anguelov2005scape} body model to generate synthetic people.
Pons-Moll et al.~\cite{pons2017clothcap} used a 3D scanner to automatically capture real clothing and estimate body shape and pose.
Compared to graphics models, image-based generative models are more computationally efficient.
Jetchev and Bergmann~\cite{jetchev2017conditional} proposed a conditional analogy GAN to swap fashion articles, without other descriptive person representation.
They didn't take pose variant into consideration, and during inference, they required the paired images of in-shop clothes and a wearer, which limits their practical scenarios.
The most related work is VITON~\cite{han2017viton}.
We all aim to synthesize photo-realistic image directly from 2D images. 
VITON addressed this problem with a coarse-to-fine framework and expected to capture the cloth deformation by a shape context TPS transoformation.
We propose an alignment network and a single pass generative framework, which preserving the characteristics of in-shop clothes.

\begin{figure}[t]
\centering
\includegraphics[height=8cm]{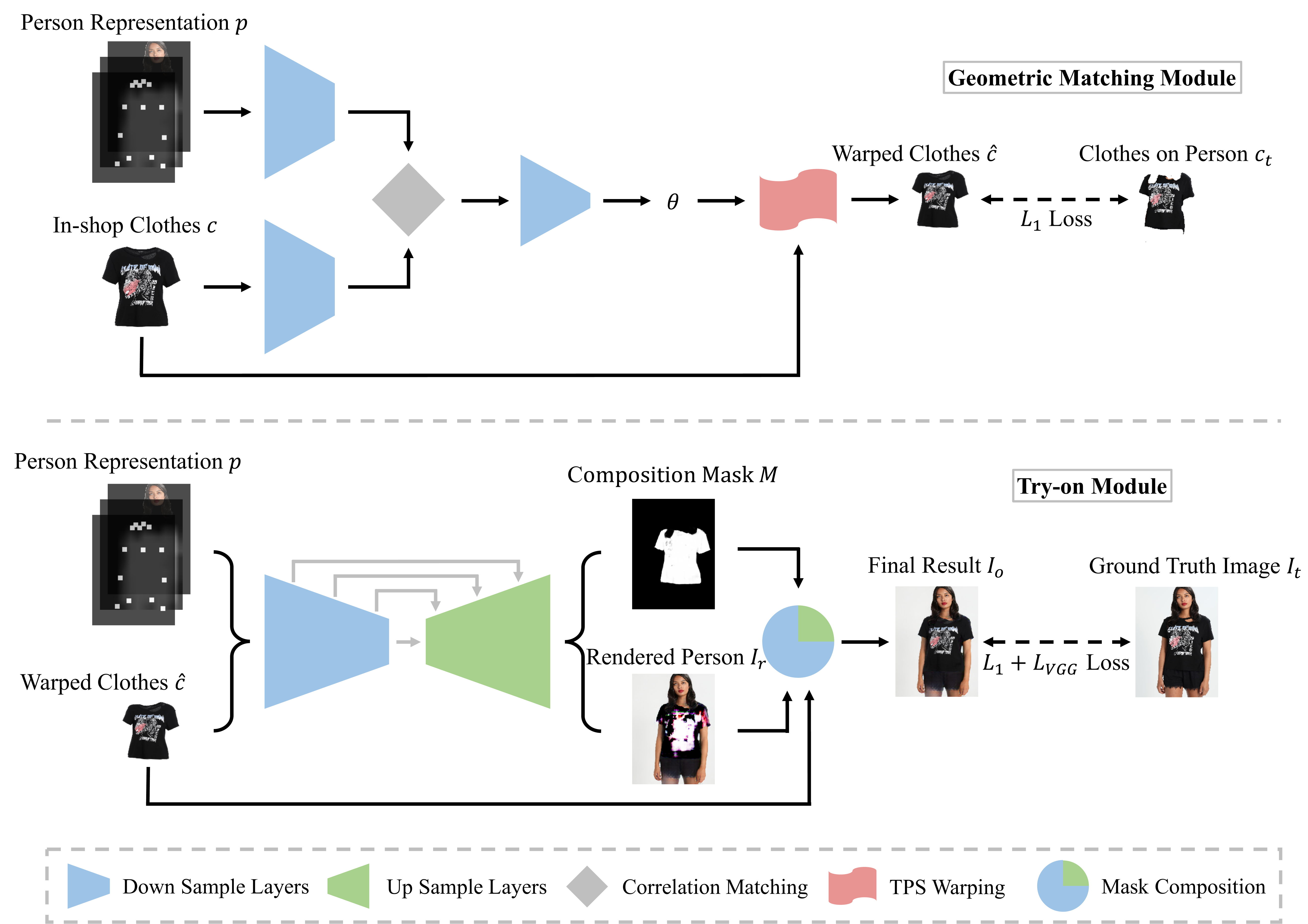}
\caption{An overview of our CP-VTON, containing two main modules. 
(a) Geometric Matching Module: the in-shop clothes $c$ and input image representation $p$ are aligned via a learnable matching module.
(b) Try-On Module: it generates a composition mask $M$ and a rendered person $I_{r}$. The final results $I_{o}$ is composed by warped clothes $\hat{c}$ and the rendered person $I_{r}$ with the composition mask $M$.}
\label{fig:framework}
\end{figure}

%
%

\section{Characteristic-Preserving Virtual Try-On Network}
\label{sections:CP-VITON}

We address the task of image-based virtual try-on as a conditional image generation problem. 
Generally, given a reference image $I_{i}$ of a person wearing in clothes $c_{i}$ and a target clothes $c$, the goal of CP-VTON is to synthesize a new image $I_{o}$ of the wearer in the new cloth $c_{o}$, in which the body shape and pose of $I_{i}$ are retained, the characteristics of target clothes $c$ are reserved and the effects of the old clothes $c_{i}$ are eliminated.

Training with sample triplets $(I_{i}, c, I_{t})$ where $I_{t}$ is the ground truth of $I_{o}$ and $c$ is coupled with $I_{t}$ wearing in clothes $c_{t}$, is straightforward but undesirable in practice.
Because these triplets are difficult to collect. 
It is easier if $I_{i}$ is same as $I_{t}$, which means that ${c, I_{t}}$ pairs are enough.
These paris are in abundance from shopping websites.
But directly training on $(I_{t}, c, I_{t})$ harms the model generalization ability at testing phase when only decoupled inputs $(I_{i}, c)$ are available.
Prior work~\cite{han2017viton} addressed this dilemma by constructing a clothing-agnostic person representation $p$ to eliminate the effects of source clothing item $c_{i}$. 
With $(I_{t}, c, I_{t})$ transformed into a new triplet form $(p, c, I_{t})$, training and testing phase are unified. 
We adopted this representation in our method and further enhance it by eliminating less information from reference person image. 
Details are described in Sec.~\ref{sections:cloth_agnostic}.
One of the challenges of image-based virtual try-on lies in the large spatial misalignment between in-shop clothing item and wearer's body. 
Existing network architectures for conditional image generation (e.g. FCN~\cite{long2015fully}, UNet~\cite{ronneberger2015u}, ResNet~\cite{he2016deep}) lack the ability to handle large spatial deformation, leading to blurry try-on results. 
We proposed a Geometric Matching Module (GMM) to explicitly align the input clothes $c$ with aforementioned person representation $p$ and produce a warped clothes image $\hat{c}$. 
GMM is a end-to-end neural network directly trained using pixel-wise L1 loss. Sec.~\ref{sections:gmm} gives the details.
Sec.~\ref{sections:try_on_module} completes our virtual try-on pipeline with a characteristic-preserving Try-On Module.
The Try-On module synthesizes final try-on results $I_{o}$ by fusing the warped clothes $\hat{c}$ and the rendered person image $I_{r}$. 
The overall pipeline is depicted in Fig.~\ref{fig:framework}.

\subsection{Person Representation}
\label{sections:cloth_agnostic}
The original cloth-agnostic person representation~\cite{han2017viton} aims at leaving out the effects of old clothes $c_{i}$ like its color, texture and shape, while preserving information of input person $I_{i}$ as much as possible, including the person's face, hair, body shape and pose. 
It contains three components:
\begin{itemize}
    \item Pose heatmap: an 18-channel feature map with each channel corresponding to one human pose keypoint, drawn as an $11 \times 11$ white rectangle.
    \item Body shape: a 1-channel feature map of a blurred binary mask that roughly covering different parts of human body.
    \item Reserved regions: an RGB image that contains the reserved regions to maintain the identity of a person, including face and hair.
\end{itemize}
These feature maps are all scaled to a fixed resolution $256 \times 192$ and concatenated together to form the cloth-agnostic person representation map $p$ of $k$ channels, where $k = 18 + 1 + 3 = 22$. 
We also utilize this representation in both our matching module and try-on module.

\subsection{Geometric Matching Module }
\label{sections:gmm}
The classical approach for the geometry estimation task of image matching consists of three stages: 
(1) local descriptors (e.g. shape context~\cite{belongie2002shape}, SIFT~\cite{lowe2004distinctive} ) are extracted from both input images,
(2) the descriptors are matched across images form a set of tentative correspondences,
(3) these correspondences are used to robustly estimate the parameters of geometric model using RANSAC~\cite{fischler1987random} or Hough voting~\cite{lamdan1988object,lowe2004distinctive}.

Rocco et al.~\cite{rocco2017convolutional} mimics this process using differentiable modules so that it can be trainable end-to-end for geometry estimation tasks.
Inspired by this work, we design a new Geometric Matching Module (GMM) to transform the target clothes $c$ into warped clothes $\hat{c}$ which is roughly aligned with input person representation $p$. 
As illustrated in Fig.~\ref{fig:framework}, our GMM consists of four parts: 
(1) two networks for extracting high-level features of $p$ and $c$ respectively.
(2) a correlation layer to combine two features into a single tensor as input to the regressor network.
(3) the regression network for predicting the spatial transformation parameters $\theta$.
(4) a Thin-Plate Spline (TPS) transformation module $T$ for warping an image into the output $\hat{c} = T_{\theta}(c)$.
The pipeline is end-to-end learnable and trained with sample triplets $(p, c, c_{t})$, under the pixel-wise L1 loss between the warped result $\hat{c}$ and ground truth $c_{t}$, where $c_t$ is the clothes worn on the target person in $I_{t}$:
\begin{align}
	\mathcal{L}_{GMM}(\theta) = ||\hat{c} - c_{t}||_1 = ||T_{\theta}(c) - c_{t}||_1
\label{formulas:GMM_Loss}
\end{align}

The key differences between our approach and Rocco et al.~\cite{rocco2017convolutional} are three-fold. 
First, we trained from scratch rather than using a pretrained VGG network. 
Second, our training ground truths are acquired from wearer's real clothes rather than synthesized from simulated warping. 
Most importantly, our GMM is directly supervised under pixel-wise L1 loss between warping outputs and ground truth.

\subsection{Try-on Module}
\label{sections:try_on_module}
Now that the warped clothes $\hat{c}$ is roughly aligned with the body shape of the target person, the goal of our Try-On module is to fuse $\hat{c}$ with the target person and for synthesizing the final try-on result.

One straightforward solution is directly pasting $\hat{c}$ onto target person image $I_{t}$. 
It has the advantage that the characteristics of warped clothes are fully preserved, but leads to an unnatural appearance at the boundary regions of clothes and undesirable occlusion of some body parts (e.g. hair, arms).
Another solution widely adopted in conditional image generation is translating inputs to outputs by a single forward pass of some encoder-decoder networks, such as UNet~\cite{ronneberger2015u}, which is desirable for rendering seamless smooth images. 
However, It is impossible to perfectly align clothes with target body shape. 
Lacking explicit spatial deformation ability, even minor misalignment could make the UNet-rendered output blurry.

Our Try-On Module aims to combine the advantages of both approaches above. 
As illustrated in Fig.~\ref{fig:framework}, given a concatenated input of person representation $p$ and the warped clothes $\hat{c}$, UNet simultaneously renders a person image $I_{r}$ and predicts a composition mask $M$. 
The rendered person $I_{r}$ and the warped clothes $\hat{c}$ are then fused together using the composition mask $M$ to synthesize the final try-on result $I_{o}$:
\begin{align}
	I_{o} = M \odot \hat{c} + (1-M) \odot I_{r}
\end{align}
where {\it $\odot$} represents element-wise matrix multiplication.

At training phase, given the sample triples $(p, c, I_{t})$, the goal of Try-On Module is to minimize the discrepancy between output $I_{o}$ and ground truth $I_{t}$. 
We adopted the widely used strategy in conditional image generation problem that using a combination of L1 loss and VGG perceptual loss \cite{johnson2016perceptual}, where the VGG perceptual loss is defined as follows:
\begin{align}
	\mathcal{L}_{\text{VGG}}(I_{o}, I_{t}) = \sum_{i=1}^{5} \lambda_{i} \left\| \phi_{i}(I_{o}) - \phi_{i}(I_{t}) \right\|_{1}
\end{align}
where {\it $\phi_{i}(I)$} denotes the feature map of image {\it $I$} of the {\it $i$-th} layer in the visual perception network {\it $\phi$}, which is a VGG19 \cite{simonyan2014very} pre-trained on ImageNet.
The layer {\it $i \geq 1$} stands for 'conv1\_2', 'conv2\_2',  'conv3\_2', 'conv4\_2', 'conv5\_2', respectively.

Towards our goal of characteristic-preserving, we bias the composition mask $M$ to select warped clothes as much as possible by applying a L1 regularization $||1 - M||_1$ on $M$. 
The overall loss function for Try-On Module (TOM) is:
\begin{align}
\label{formulas:tom_loss}
	\mathcal{L}_{\text{TOM}} = \lambda_{L1}||I_{o} - I_{t}||_1 + \lambda_{vgg} \mathcal{L}_{\text{VGG}}(\hat{I}, I) + \lambda_{mask} ||1 - M||_1.
\end{align}

%
%
\section{Experiments and Analysis}
\subsection{Dataset}
We conduct our all experiments on the datasets collected by Han et al.~\cite{han2017viton}.
It contains around 19,000 front-view woman and top clothing image pairs.
There are 16253 cleaned pairs, which are split into a training set and a validation set with 14221 and 2032 pairs, respectively.
We rearrange the images in the validation set into unpaired pairs as the testing set.

\subsection{Quantitative Evaluation}
\label{sections:quatitative}
\begin{figure*}
\centering
\includegraphics[scale=0.3]{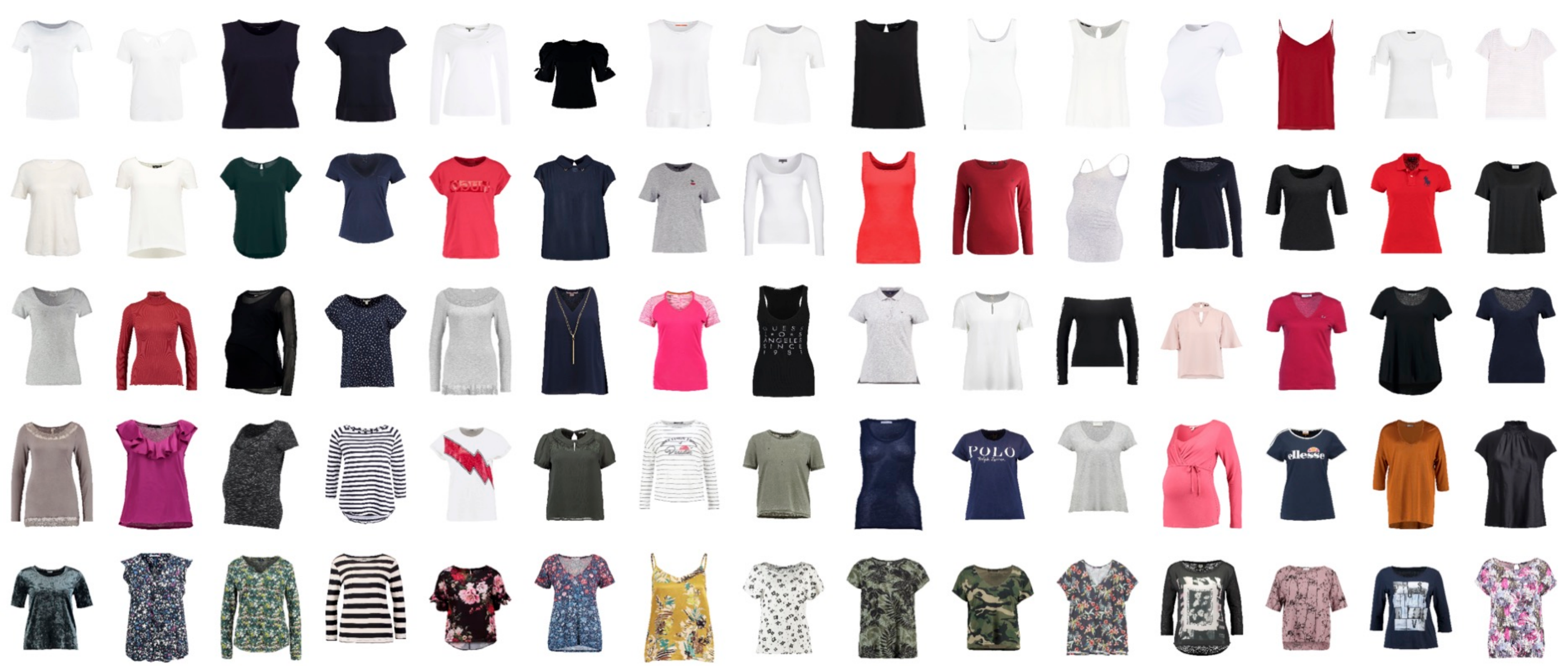}
\caption{From top to bottom, the TV norm values are increasing. Each line shows some clothes in the same level.}
\label{fig:tv_groups}
\end{figure*}

We evaluate the quantitative performance of different virtual try-on methods via a human subjective perceptual study.
Inception Score (IS) ~\cite{salimans2016improved} is usually used as to quantitatively evaluate the image synthesis quality, 
but not suitable for evaluating this task for that it cannot reflect whether the details are preserved as described in~\cite{han2017viton}.
We focus on the clothes with rich details since we are interested in characteristic-preservation, instead of evaluating on the whole testing set.
For simplicity, we measure the detail richness of a clothing image by its total variation (TV) norm. 
It is appropriate for this dataset since the background is in pure color and the TV norm is only contributed by clothes itself, as illustrated in Fig.~\ref{fig:tv_groups}.
We extracted 50 testing pairs with largest clothing TV norm named as {\bf LARGE} to evaluate characteristic-preservation of our methods, 
and 50 pairs with smallest TV norm named as {\bf SMALL} to ensure that our methods perform at least as good as previous state-of-the-art methods in simpler cases.

We conducted pairwise A/B tests on Amazon Mechanical Turk (AMT) platform. 
Specifically, given a person image and a target clothing image, the worker is asked to select the image which is more realistic and preserves more details of the target clothes between two virtual try-on results from different methods.
There is no time limited for these jobs, and each job is assigned to 4 different workers.
Human evaluation metric is computed in the same way as in \cite{han2017viton}.

\subsection{Implementation Details}

\begin{figure*}[t!]
\centering
\includegraphics[height=9cm]{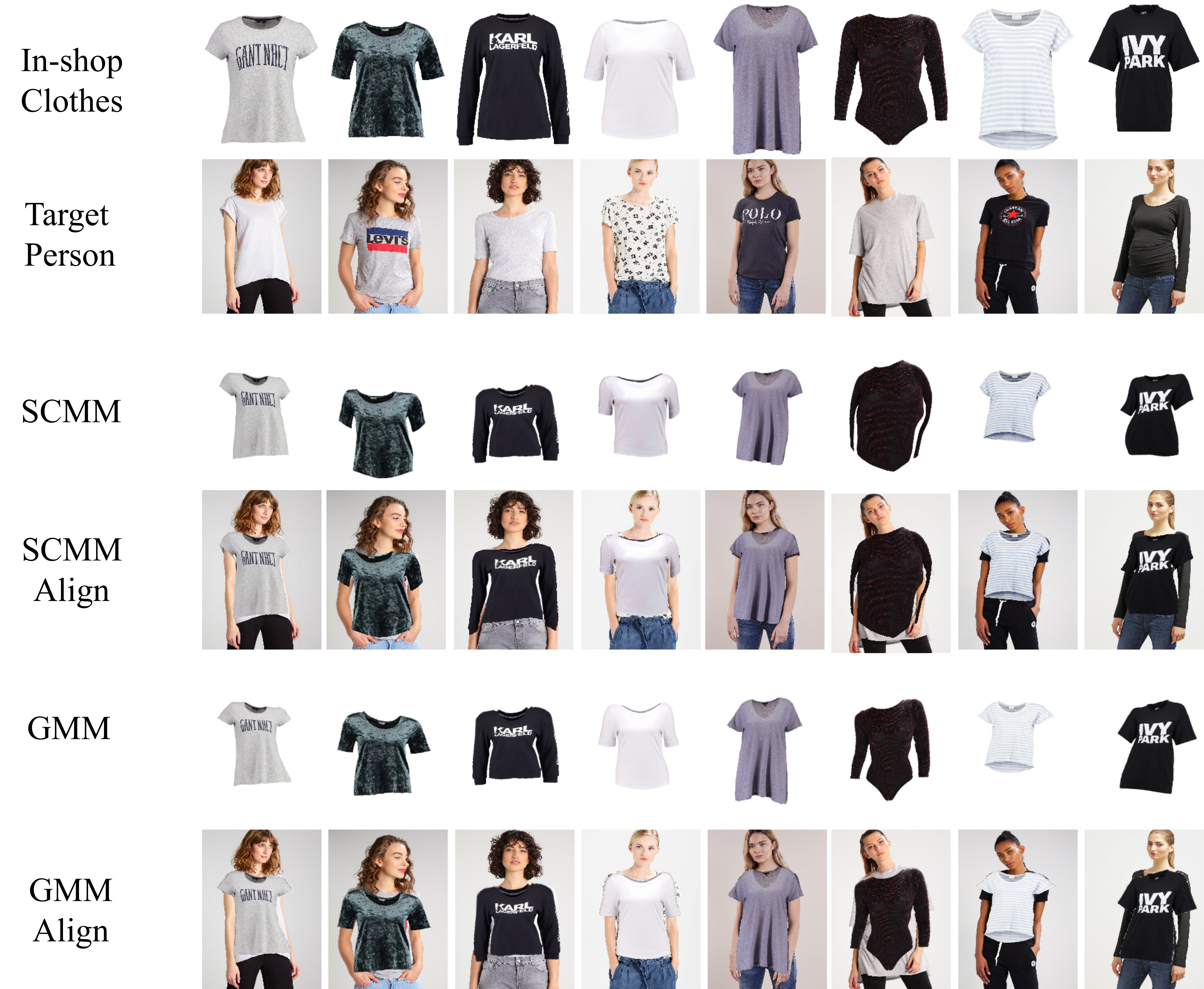}
\caption{Matching results of SCMM and GMM. Warped clothes are directly pasted onto target persons for visual checking. Our method is comparable with SCMM and produces less weird results.}
\label{fig:gmm_vs_scmm_align}
\end{figure*}

\noindent\textbf{Training Setup}
In all experiments, we use $\lambda_{L1} = \lambda_{vgg} = 1$. When composition mask is used, we set $\lambda_{mask} = 1$. 
We trained both Geometric Matching Module and Try-on Module for 200K steps with batch size 4.  
We use Adam \cite{kinga2015method} optimizer with $\beta_1 = 0.5$ and $\beta_2 = 0.999$. 
Learning rate is first fixed at 0.0001 for 100K steps and then linearly decays to zero for the remaining steps. 
All input images are resized to $256 \times 192$ and the output images have the same resolution.

\noindent\textbf{Geometric Matching Module}
Feature extraction networks for person representation and clothes have the similar structure, containing four 2-strided down-sampling convolutional layers, succeeded by two 1-strided ones, their numbers of filters being 64, 128, 256, 512, 512, respectively. 
The only difference is the number of input channels. 
Regression network contains two 2-strided convolutional layers, two 1-strided ones and one fully-connected output layer. 
The numbers of filters are 512, 256, 128, 64. 
The fully-connected layer predicts the x- and y-coordinate offsets of TPS anchor points, thus has an output size of $2 \times 5 \times 5 = 50$. 

\noindent\textbf{Try-On Module}
We use a 12-layer UNet with six 2-strided down-sampling convolutional layers and six up-sampling layers. 
To alleviate so-called ``checkerboard artifacts", we replace 2-strided deconvolutional layers normally used for up-sampling with the combination of nearest-neighbor interpolation layers and 1-strided convolutional layers, as suggested by \cite{odena2016deconvolution}. 
The numbers of filters for down-sampling convolutional layers are 64, 128, 256, 512, 512, 512. 
The numbers of filters for up-sampling convolutional layers are 512, 512, 256, 128, 64, 4. 
Each convolutional layer is followed by an Instance Normalization layer \cite{ulyanov2017improved} and Leaky ReLU~\cite{maas2013rectifier}, of which the slope is set to 0.2.

\subsection{Comparison of Warping Results}
Shape Context Matching Module (SCMM) uses hand-crafted descriptors and explicitly computes their correspondences using an iterative algorithm, which is time-consumed, while GMM runs much faster. In average, processing a sample pair takes GMM 0.06s on GPU, 0.52s on CPU, and takes SCMM 2.01s on CPU.

\noindent\textbf{Qualitative results}
Fig.~\ref{fig:gmm_vs_scmm_align} demonstrates a qualitative comparison of SCMM and GMM.
It shows that both modules are able to roughly align clothes with target person pose. 
However, SCMM tends to overly shrink a long sleeve into a ``thin band", as shown in the $6$-th column in Fig.~\ref{fig:gmm_vs_scmm_align}. 
This is because SCMM merely relies on matched shape context descriptors on the boundary of cloths shape, while ignores the internal structures. 
Once there exist incorrect correspondences of descriptors, the warping results will be weird. 
In contrast, GMM takes full advantages of the learned rich representation of clothes and person images to determinate TPS transformation parameters and more robust for large shape differences.

\noindent\textbf{Quantitative results}
It is difficult to evaluate directly the quantitative performance of matching modules due to the lack of ground truth in the testing phase. 
Nevertheless, we can simply paste the warped clothes onto the original person image as a non-parametric warped synthesis method in \cite{han2017viton}.
We conduct a perceptual user study following the protocol described in Sec.~\ref{sections:quatitative}, for these two warped synthesis methods.
The synthesized by GMM are rated more realistic in 49.5\% and 42.0\% for {\bf LARGE} and {\bf SMALL}, which indicates that GMM is comparable to SCMM for shape alignment.

\subsection{Comparison of Try-on Results}
\label{sections:comparison_try_on}

\begin{figure}[t]
\centering
\includegraphics[height=6cm]{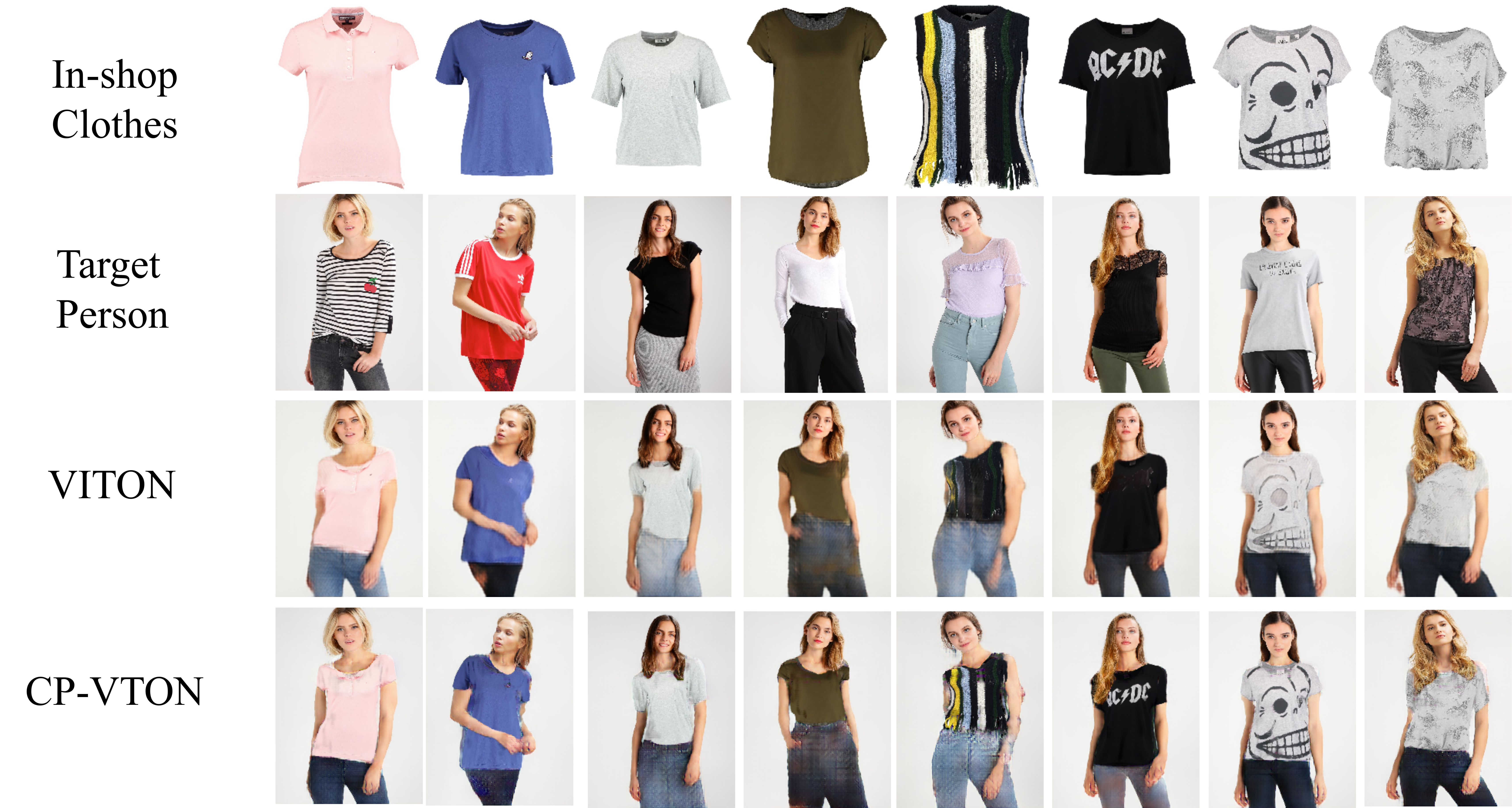}
\caption{Qualitative comparisons of VITON and CP-VTON. Our CP-VTON successfully preserve key details of in-shop clothes.}\vspace{-4mm}
\label{fig:viton_vs_ours}
\end{figure}

\noindent\textbf{Qualitative results}
Fig.~\ref{fig:framework} shows that our pipeline performs roughly the same as VITON  when the patterns of target clothes are simpler. 
However, our pipeline preserves sharp and intact characteristic on clothes with rich details (e.g. texture, logo, embroidery) while VITON produces blurry results. 

\begin{figure*}[t]
\centering
\includegraphics[height=2cm]{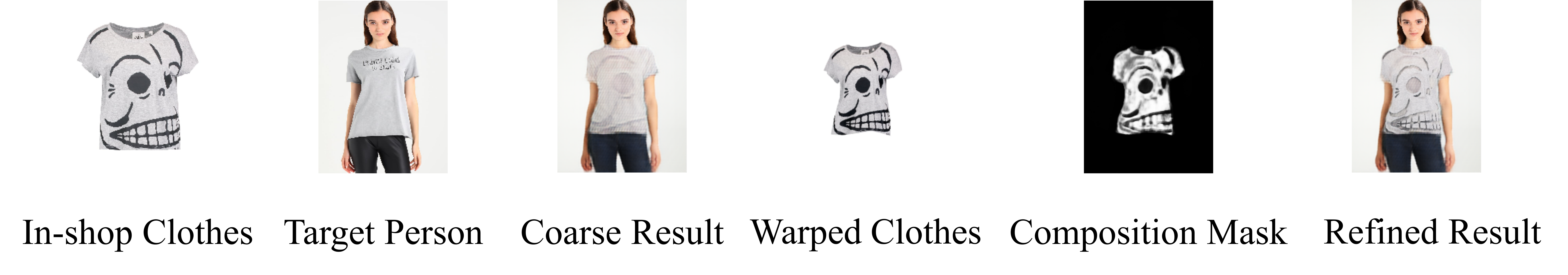}
\caption{An example of VITON stage \uppercase\expandafter{\romannumeral2}. The composition mask tends to ignore the details of coarsely aligned clothes.}
\label{fig:viton_results}
\end{figure*}

We argue that the failure of VITON lies in its coarse-to-fine strategy and the imperfect matching module. 
Precisely, VITON learns to synthesis a coarse person image at first, then to align the clothes with target person with shape context matching, then to produce a composition mask for fusing UNet rendered person with warped clothes and finally producing a refined result. 
After extensive training, the rendered person image has already a small VGG perceptual loss with respect to ground truth. 
On the other hand, the imperfect matching module introduces unavoidable minor misalignment between the warped clothes and ground truth, making the warped clothes unfavorable to perceptual loss. 
Taken together, when further refined by truncated perceptual loss, the composition mask will be biased towards selecting rendered person image rather than warped clothes, despite the regularization of the composition mask(Eq.~\ref{formulas:tom_loss}).  
The VITON's ``ragged" masks shown in Fig.~\ref{fig:viton_results} confirm this argument.

\setlength{\tabcolsep}{4pt}
\begin{table}[t]
\begin{center}
\caption{Results of pairwise comparisons of images synthesized with {\bf LARGE} and {\bf SMALL} clothes by different models. Each column compares our approach with one of the baselines. Higher is better. The random chance is at 50\%.}\vspace{-3mm}
\label{table:comparison}
\begin{tabular}{ccccc}
\hline\noalign{\smallskip}
Data & VITON & CP-VTON(w/o mask) & CP-VTON(w/o L1 Loss)\\
\noalign{\smallskip}
\hline
\noalign{\smallskip}
{\bf LARGE} & 67.5\% & 72.5\% & 84.5\%\\
{\bf SMALL} & 55.0\%  & 42\% & 38.5\%\\
\hline
\end{tabular}
\end{center}
\end{table}
\setlength{\tabcolsep}{1.4pt}

Our pipeline doesn't address the aforementioned issue by improving matching results, but rather sidesteps it by simultaneously learning to produce a UNet rendered person image and a composition mask. 
Before the rendered person image becomes favorable to loss function, the central clothing region of composition mask is biased towards warped clothes because it agrees more with ground truth in the early training stage. 
It is now the warped clothes rather than the rendered person image that takes the early advantage in the competition of mask selection.
After that, the UNet learns to adaptively expose regions where UNet rendering is more suitable than directly pasting. 
Once the regions of hair and arms are exposed, rendered and seamlessly fused with warped clothes.

\noindent\textbf{Quantitative results}
The first column of Table~\ref{table:comparison} shows that our pipeline surpasses VITON in the preserving the details of clothes using identical person representation.
According to the table, our approach performs better than other methods, when dealing with rich details clothes.

\subsection{Discussion and Ablation Studies}

\subsubsection{Effects of composition mask}
To empirically justify the design of composition mask and mask L1 regularization (Eq.~\ref{formulas:tom_loss}) in our pipeline, we compare it with two variants for ablation studies: 
(1): mask composition is also removed and the final results are directly rendered by UNet as CP-VTON(w/o mask).
(2): the mask composition is used but the mask L1 regularization is removed as CP-VTON(w/o L1 Loss);

As shown in Fig.~\ref{fig:viton_results}, even though the warped clothes are roughly aligned with target person, CP-VTON(w/o mask) still loses characteristic details and produces blurry results. This verifies that encoder-decoder network architecture like UNet fails to handle even minor spatial deformation.

\begin{figure}[t]
\centering
\includegraphics[height=7cm]{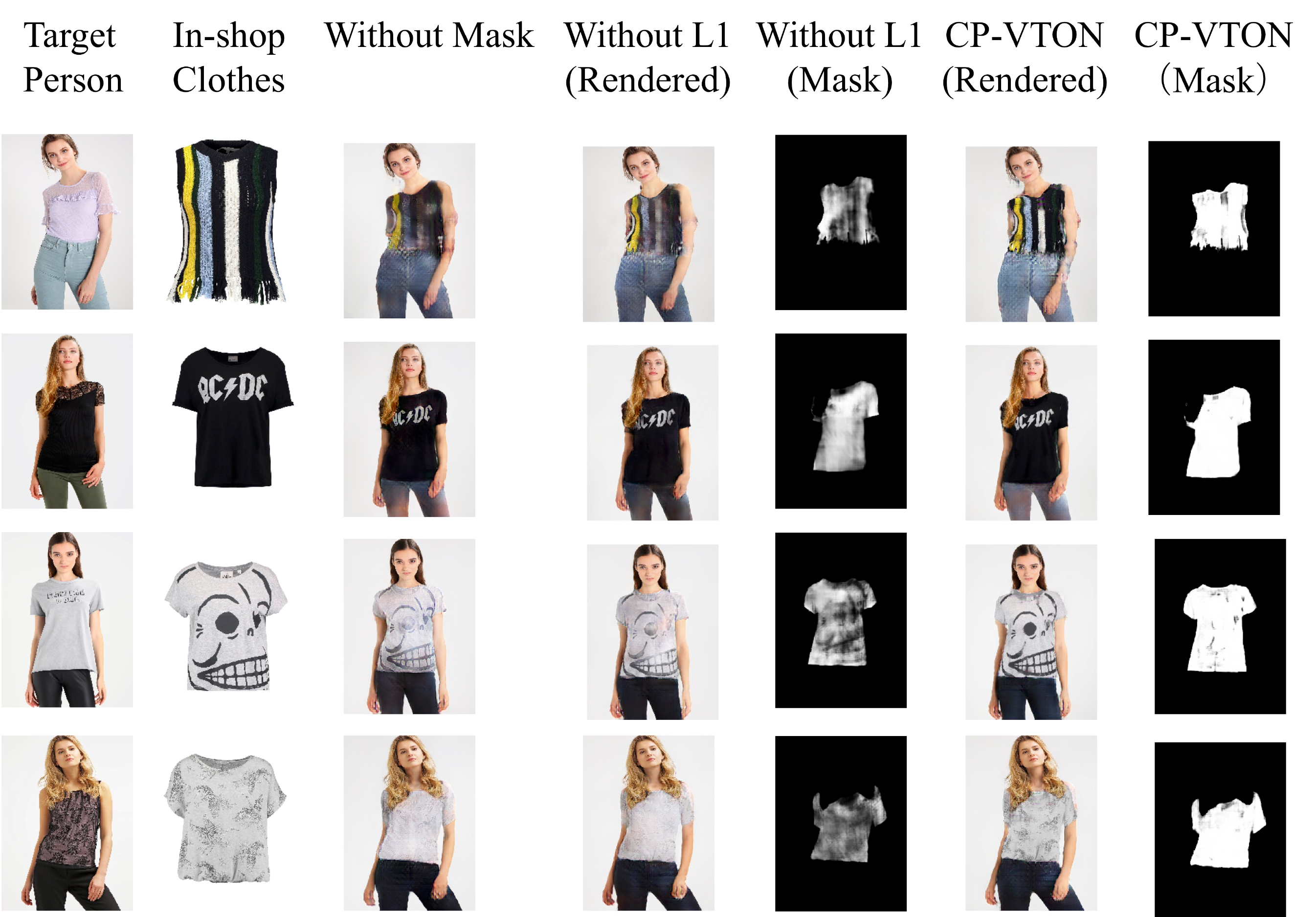}
\caption{Ablation studies on composition mask and mask L1 loss. Without mask composition, UNet cannot handle well even minor misalignment and produces undesirable try-on results. Without L1 regularization on mask, it tends to select UNet-rendered person, leading to blurry results as well.}\vspace{-4mm}
\label{fig:our_ablation}
\end{figure}

Though integrated with mask composition, CP-VTON(no L1) performs as poorly as variant CP-VTON(w/o mask. Fig.~\ref{fig:our_ablation} shows that composition mask tends to select rendered person image without L1 regularization. This verifies that even minor misalignment introduces large perceptual disagreement between warped clothes and ground truth.

\subsubsection{Robustness against minor misalignment}
In Sec.~\ref{sections:comparison_try_on} we argue that VITON is vulnerable to minor misalignment due to its coarse-to-fine strategy, while our pipeline sidesteps imperfect alignment by simultaneously producing rendered person and composition mask. 
This is further clarified below in a controlled condition with simulated warped clothes.

Specifically, rather than real warped clothes produced by matching module, we use the wore clothes collected from person images to simulate perfect alignment results. We then train VITON stage \uppercase\expandafter{\romannumeral2}, our proposed variant  CP-VTON(w/o mask) and our pipeline. 
For VITON stage \uppercase\expandafter{\romannumeral2}, we synthesize coarse person image with its source code and released model checkpoint.

It is predictable that with this ``perfect matching module", all the three methods could achieve excellent performance in training and validation phase, where input samples are paired. 
Next is the interesting part: what if the perfect alignment is randomly perturbed within a range of $N$ pixels, to simulate an imperfect matching module? 
With the perturbation getting greater ($N = 0, 5, 10, 15, 20$) , how fast will the try-on performance decay?

\begin{figure}[t]
\centering
\includegraphics[height=5cm]{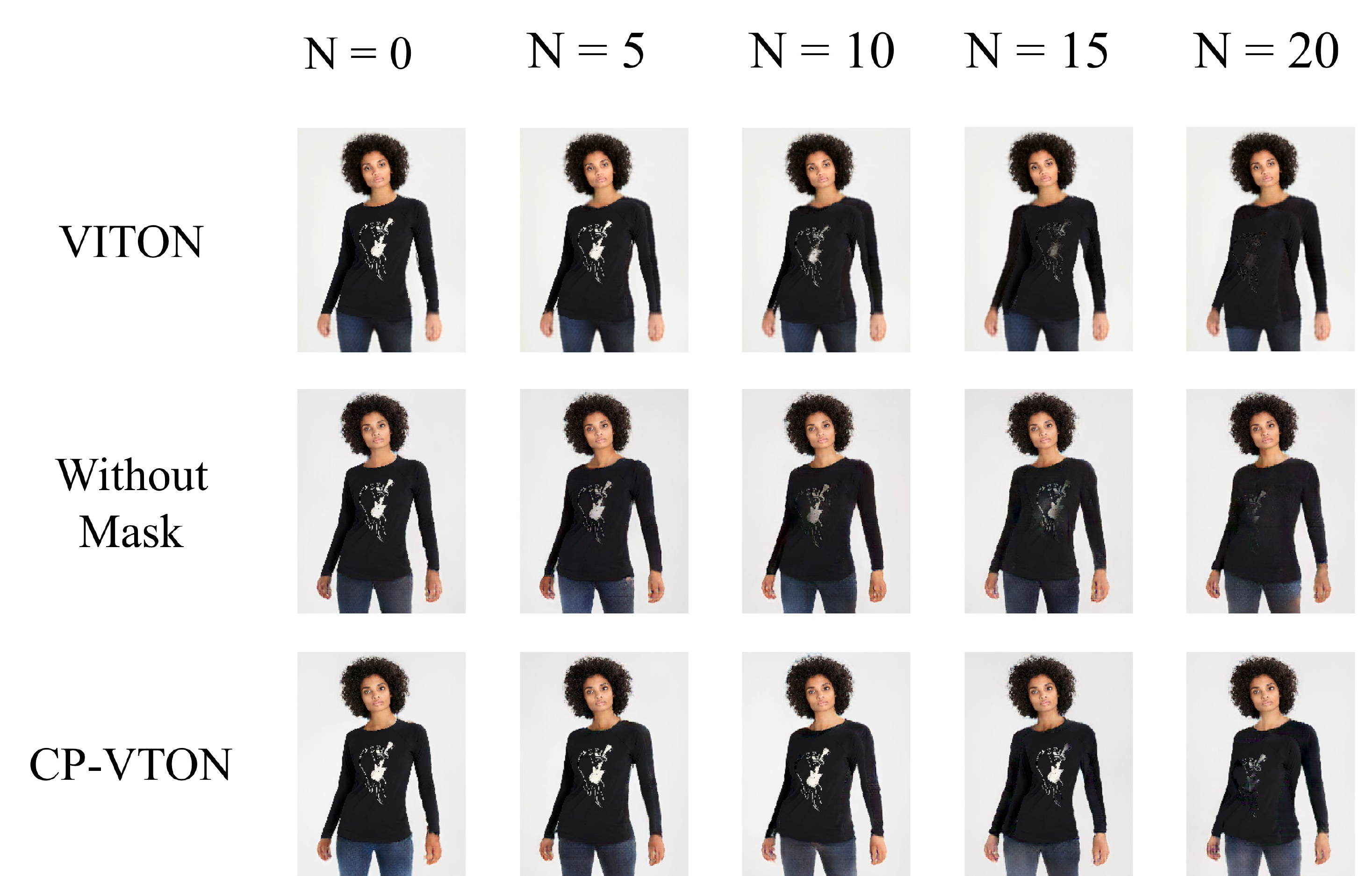}
\caption{Comparisons on the robustness of three methods against minor misalignment simulated by random shift within radius $N$. As $N$ increasing, results of CP-VTON decays more slightly than other methods.}
\label{fig:simulate_misalign}
\end{figure}

These questions are answered in Fig. ~\ref{fig:simulate_misalign}. As we applying greater perturbation, the performance of both VITON stage \uppercase\expandafter{\romannumeral2} and CP-VTON(w/o mask)  decays quickly. 
In contrast, our pipeline shows robustness against perturbation and manages to preserve detailed characteristic.

\subsubsection{Failure cases}

\begin{figure}[t]
\centering
\includegraphics[height=1.6cm]{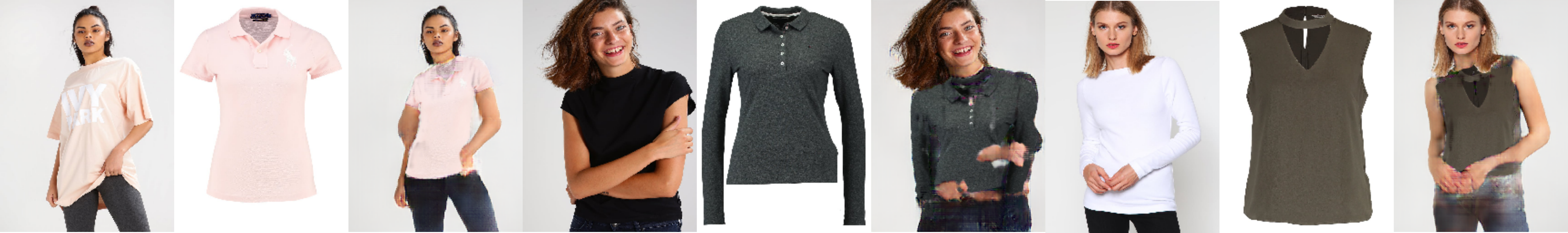}
\caption{Some failure cases of our CP-VTON.}
\label{fig:failure_case}\vspace{-6mm}
\end{figure}

Fig. ~\ref{fig:failure_case} shows three failure cases of our CP-VTON method caused by (1) improperly preserved shape information of old clothes, (2) rare poses and (3) inner side of the clothes undistinguishable from the outer side, respectively.

%
%
\section{Conclusions}

In this paper, we propose a fully learnable image-based virtual try-on pipeline towards the characteristic-preserving image generation, named as CP-VTON, including a new geometric matching module and a try-on module with the new merging strategy. 
The geometric matching module aims at aligning in-shop clothes and target person body with large spatial displacement. 
Given aligned clothes, the try-on module learns to preserve well the detailed characteristic of clothes. 
Extensive experiments show the overall CP-VTON pipeline produces high-fidelity virtual try-on results that retain well key characteristics of in-shop clothes. 
Our CP-VTON achieves state-of-the-art performance on the dataset collected by Han et al.~\cite{han2017viton} both qualitatively and quantitatively.
%
%
%
\bibliographystyle{splncs04}
\bibliography{eccv2018-cpvton}

\end{document}